\documentclass[sigconf]{acmart}
\usepackage{booktabs} 
\usepackage{multirow} 
\usepackage{colortbl} 
\usepackage{enumitem}
\usepackage{xcolor}  
\usepackage{svg}
\usepackage{subcaption}
\usepackage{graphicx}
\usepackage{balance}
\AtBeginDocument{%
  }

\copyrightyear{2026}
\acmYear{2026}
\setcopyright{cc}
\setcctype{by}
\acmConference[MM '26] {Proceedings of the 34th ACM International Conference on Multimedia}{November 10--14, 2026}{Rio de Janeiro, Brazil.}
\acmBooktitle{Proceedings of the 34th ACM International Conference on Multimedia (MM '26), November 10--14, 2026, Rio de Janeiro, Brazil}
\acmISBN{979-8-4007-2213-4/2026/11}
\acmDOI{10.1145/3767308.3836554}

\begin{document}
\author{Xiaowei Mao}
\authornote{Equal contribution.}
\author{Bowen Sui}
\authornotemark[1]

\author{Weijie Zhang}
\author{Yawen Yang}
\affiliation{
  \department{School of Computer Science and Technology}
  \institution{Beijing Jiaotong University}
  \city{Beijing}
  \country{China}
}
\email{maoxiaowei@bjtu.edu.cn}

\author{Shengnan Guo}
\authornote{Corresponding author.}
\affiliation{
  \department{School of Computer Science and Technology}
  \institution{Beijing Jiaotong University}
  \city{Beijing}
  \country{China}
}
\affiliation{
  \institution{Key Laboratory of Big Data \& Artificial Intelligence in Transportation, Ministry of Education}
  \city{Beijing}
  \country{China}
}
\email{guoshn@bjtu.edu.cn}

\author{Shilong Zhao}
\author{Jiaqi Lin}
\author{Tingrui Wu}
\affiliation{
  \department{School of Computer Science and Technology}
  \institution{Beijing Jiaotong University}
  \city{Beijing}
  \country{China}
}

\author{Youfang Lin}
\affiliation{
  \department{School of Computer Science and Technology}
  \institution{Beijing Jiaotong University}
  \city{Beijing}
  \country{China}
}
\affiliation{
  \institution{Beijing Key Laboratory of Traffic Data Mining and Embodied Intelligence}
  \city{Beijing}
  \country{China}
}

\author{Huaiyu Wan}
\affiliation{
  \department{School of Computer Science and Technology}
  \institution{Beijing Jiaotong University}
  \city{Beijing}
  \country{China}
}
\affiliation{
  \institution{Beijing Key Laboratory of Traffic Data Mining and Embodied Intelligence}
  \city{Beijing}
  \country{China}
}
\renewcommand{\shortauthors}{Xiaowei Mao et al.}

\title[VIBES: Efficient Far-field Anomaly Detection in Expressway Surveillance Videos]
{Zoom In, Reason Out: Efficient Far-field Anomaly Detection in Expressway Surveillance Videos via Focused VLM Reasoning Guided by Bayesian Inference}

\begin{abstract}

Expressway video anomaly detection is important for traffic safety, but remains challenging across diverse scenes, particularly for far-field vehicles with subtle abnormal motion. Vision-Language Models (VLMs) provide strong semantic reasoning capabilities, yet processing full frames can dilute evidence from distant targets and introduce substantial computational overhead. To address these challenges, we propose VIBES, an asynchronous framework that uses Bayesian inference to guide focused VLM reasoning. Specifically, an online kinematics-guided Bayesian inference module continuously estimates a context-dependent normal-motion distribution from vehicle trajectories and updates its probabilistic boundaries. Deviations from these boundaries produce asynchronous triggers that localize candidate anomalies in time and space. Instead of processing continuous full-frame video, the VLM reasons only over selected frames and localized visual regions associated with the triggers, reducing irrelevant visual content and unnecessary inference. Extensive experiments show that VIBES improves far-field anomaly detection and semantic interpretation while achieving real-time processing efficiency across diverse expressway conditions.

\end{abstract}


\maketitle

\section{Introduction}

Expressway video anomaly detection is important for traffic safety. Surveillance cameras continuously monitor road networks, where timely detection of abnormal driving behaviors, such as sudden lane deviations and illegal parking, can support rapid traffic management~\cite{santhosh2020anomaly,sultani2018real,abdalla2025video}. However, accurate and efficient anomaly detection remains challenging.

\begin{figure*}[t]
  \centering
  \includegraphics[width=\textwidth]{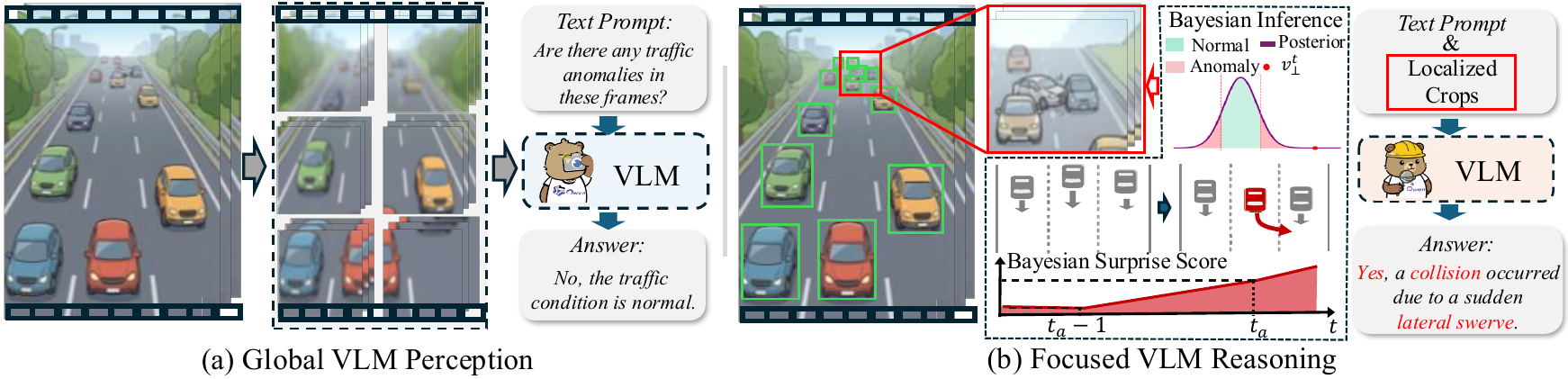}
  \caption{Comparison of expressway anomaly detection paradigms. 
  (a) Global VLM perception processes full frames, where far-field anomalous evidence can be overwhelmed by irrelevant visual content. 
  (b) Focused VLM reasoning uses Bayesian inference to localize anomalous motion and provide compact visual evidence for semantic reasoning.}
  \label{fig:intro}
\end{figure*}

First, far-field anomalies are difficult to detect because distant vehicles occupy only small image regions~\cite{akyon2022slicing,miri2025transformers}. As shown in Figure~\ref{fig:intro}, the far field often covers a large portion of the road while individual targets contain limited visual detail. Its location also varies with camera viewpoint and road geometry, while normal motion patterns change with traffic conditions. These factors make anomaly detection difficult to generalize across scenes. Existing methods model normal appearance or motion~\cite{lv2021learning,yuan2016anomaly}, analyze vehicle trajectories~\cite{zhao2019unsupervised}, or detect inconsistencies in predicted object motion~\cite{yao2022dota}. Although effective in specific settings, they can be sensitive to scene variation and typically provide limited semantic interpretation.

Recent Vision-Language Models (VLMs) support semantic reasoning for visual anomaly understanding~\cite{D-VLP, Bridging-domain}, while models such as Qwen3-VL~\cite{Qwen3-VL} improve fine-grained visual perception. However, full-frame processing remains challenging for far-field anomalies, as small anomalous regions provide limited evidence amid dominant normal and visually salient content~\cite{zhong2025focus,zhang2025cofft,miri2025transformers}. As illustrated in Figure~\ref{fig:intro}(a), this visual dilution can cause VLMs to overlook distant abnormal motion.

Second, applying VLMs continuously to surveillance streams is computationally expensive. Frequent VLM inference increases processing latency, whereas sparse frame sampling reduces computation at the risk of missing short anomalous events~\cite{ding2025videozoomer,ding2025streammind}. Therefore, effective expressway anomaly detection requires both reliable localization of subtle far-field motion and selective VLM reasoning.

To address these challenges, we propose VIBES, an asynchronous framework utilizing \underline{Vi}sion-language models guided by \underline{B}ay\underline{es}ian inference for far-field anomaly detection in expressway surveillance videos. As shown in Figure~\ref{fig:intro}, VIBES follows a {zoom-in and reason-out} pipeline.

First, we introduce a kinematics-guided Bayesian inference module for far-field motion detection. Lightweight trajectory processing decomposes vehicle motion into anomaly-related kinematic patterns. The module continuously estimates a context-dependent normal-motion distribution from history and surrounding traffic, and dynamically updates its probabilistic boundaries. When recent motion violates these boundaries, a Bayesian surprise score triggers the corresponding time and vehicle region for further analysis.

Second, instead of processing the continuous full-frame stream, VIBES selects only the temporal frames and localized regions associated with each trigger. The resulting focused visual evidence is then processed by the VLM for semantic reasoning. By restricting VLM inference to event-centered inputs, VIBES reduces irrelevant visual content and unnecessary VLM invocations while preserving the evidence required for anomaly interpretation. The main contributions are summarized as follows:

\begin{itemize}[leftmargin=*, topsep=0pt]
    \item We propose VIBES, an asynchronous framework that uses Bayesian inference to guide VLM reasoning over focused visual evidence, mitigating attention dilution for far-field anomalies.
    \item We introduce a kinematics-guided Bayesian inference module that dynamically updates normal-motion boundaries to trigger localized anomaly detection, reducing VLM invocations while maintaining accuracy.
    \item Extensive experiments demonstrate that VIBES achieves high detection accuracy and real-time efficiency across diverse expressway conditions.
\end{itemize}

\section{Related Work}

\subsection{Traffic Video Anomaly Detection}

Video anomaly detection (VAD) has evolved from visual normality modeling based on reconstruction, prediction, or memory mechanisms~\cite{hasan2016temporal,liu2018future,gong2019memorizing,park2020learning}, to weakly supervised learning from video-level labels~\cite{sultani2018real,tian2021rtfm}, and more recently to language-aided anomaly understanding~\cite{wu2024vadclip,zanella2024lavad,wu2024open,zhang2025holmesvau,ye2025vera}. Traffic-oriented methods often exploit vehicle-level motion cues: trajectory-based approaches identify unusual motion patterns~\cite{zhao2019unsupervised}, while DoTA~\cite{yao2022dota} detects inconsistencies between predicted and observed future object locations.

Despite this progress, far-field expressway anomalies remain challenging. Distant vehicles occupy small image regions, causing localized motion deviations to contribute weakly to full-frame reconstruction or prediction~\cite{liu2018future,gong2019memorizing,park2020learning}. Vehicle tracking and trajectory analysis preserve localized motion cues, but far-field targets are more sensitive to missed detections, localization noise, and track fragmentation~\cite{zhao2019unsupervised,yao2022dota}. Moreover, apparent motion varies with camera viewpoint, road geometry, and traffic conditions, complicating normal-motion modeling across scenes. Language-based VAD provides richer semantic interpretation, but reliable spatiotemporal localization of small distant anomalies remains difficult.

\subsection{Video Understanding with VLMs}

Recent VLM research on video understanding has focused on efficient long-video processing and fine-grained visual reasoning. VideoLLM-online~\cite{chen2024videollm}, MovieChat~\cite{song2024moviechat}, and StreamMind~\cite{ding2025streammind} reduce streaming or long-video cost through online processing, memory, or event-gated computation. Frame-selection methods such as M-LLM~\cite{hu2025m} and MDP$^3$~\cite{sun2025mdp3} retain informative frames, while LongVU~\cite{shen2025longvu} and APVR~\cite{gao2026apvr} compress or retrieve visual information under limited budgets. Fine-grained reasoning methods further improve access to localized evidence: V$^*$~\cite{wu2024vstar} performs guided visual search, FOCUS~\cite{zhong2025focus} identifies relevant image regions, and DeepScan~\cite{li2026deepscan} progressively searches and refines local visual evidence.

However, these approaches do not directly address the spatiotemporal localization of subtle far-field motion. Frame selectors operate primarily at the frame level, where a small anomalous region may contribute little to the overall relevance score, and temporal search over long videos remains challenging~\cite{ye2025temporalsearch}. Memory, compression, and retrieval reduce visual cost but do not explicitly identify the distant region exhibiting abnormal motion. Fine-grained reasoning methods improve local inspection, but iterative region search can introduce additional inference overhead when applied continuously. Thus, efficient localization of small, motion-related anomalies remains challenging under varying viewpoints and traffic conditions.

\section{Preliminaries and Problem Statement}
\label{sec:preliminaries}

This section introduces the notation for tracked vehicle motion and defines the task addressed by VIBES.

\noindent\textbf{Definition 1 (Expressway surveillance video).}
An expressway surveillance video is denoted by
$\mathcal{V}=\{\mathbf{I}_1,\mathbf{I}_2,\ldots,\mathbf{I}_T\}$, where
$\mathbf{I}_t\in\mathbb{R}^{H\times W\times 3}$ is the frame at time
$t\in\{1,\ldots,T\}$, and $H$ and $W$ denote the frame height and width.

\noindent\textbf{Definition 2 (Tracked vehicle trajectory).}
The trajectory of vehicle $i$ is
$\mathcal{T}_i=\{(t_r,\mathbf{b}_i^{t_r})\}_{r=1}^{n_i}$, where
$1\le t_1<t_2<\cdots<t_{n_i}\le T$.
At the $r$-th valid observation,
$\mathbf{b}_i^{t_r}=(c_{x,i}^{t_r},c_{y,i}^{t_r},b_{w,i}^{t_r},b_{h,i}^{t_r})
\in\mathbb{R}^4$ denotes its bounding box, with center
$(c_{x,i}^{t_r},c_{y,i}^{t_r})$, width $b_{w,i}^{t_r}$, and height
$b_{h,i}^{t_r}$. The observation times need not be consecutive; $r$
indexes valid observations, while $t_r$ denotes the corresponding frame.

\noindent\textbf{Problem statement (Anomaly detection and explanation).}
Given $\mathcal{V}$ and a task prompt $\mathcal{P}$, VIBES detects anomalous
vehicle motion and generates an event description:
\begin{equation}
\mathcal{Y}=F(\mathcal{V},\mathcal{P}),
\qquad
\mathcal{Y}=(y_1,\ldots,y_{L_y})
\label{eq:problem}
\end{equation}
The output reports the anomaly time $t_a$, event category, involved vehicles,
and the observed cause or consequence.

\begin{figure*}[t]
  \centering
  \includegraphics[width=.95\textwidth]{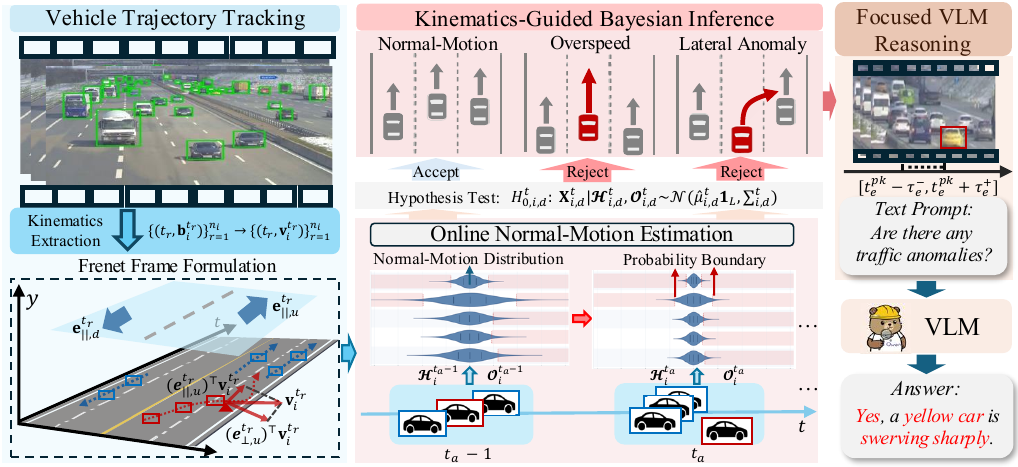}
\caption{The architecture of VIBES. Left: Trajectory tracking extracts vehicle kinematics. Middle: Kinematics-guided Bayesian inference dynamically updates probabilistic boundaries to detect anomalies. Right: Focused VLM reasoning utilizes these triggers to extract localized visual prompts, enabling accurate explanation of the anomalous events.}
  \label{fig:framework}
\end{figure*}

\section{Methods}

\subsection{Overview}
\label{sec:overview}

As illustrated in Fig.~\ref{fig:framework}, VIBES follows an asynchronous
{zoom-in and reason-out} pipeline. A lightweight kinematic module continuously
tracks vehicle motion and estimates a context-dependent normal-motion distribution.
Deviations beyond the corresponding probability boundaries generate kinematic
triggers, which are grouped into event proposals for spatiotemporal localization.
The selected frames and localized regions are then provided to the VLM for semantic
reasoning. Formally,
\begin{equation}
\mathcal{A}=\mathcal{F}_{\mathrm{kin}}(\mathcal{V}),
\qquad
\mathcal{X}^{\star}=\mathcal{F}_{\mathrm{loc}}(\mathcal{V},\mathcal{A}),
\qquad
\mathcal{Y}=\mathcal{F}_{\mathrm{vlm}}(\mathcal{X}^{\star},\mathcal{P}),
\label{eq:overview_pipeline}
\end{equation}
where $\mathcal{F}_{\mathrm{kin}}$ produces the kinematic trigger set
$\mathcal{A}$, $\mathcal{F}_{\mathrm{loc}}$ extracts focused spatiotemporal
evidence $\mathcal{X}^{\star}$, and $\mathcal{F}_{\mathrm{vlm}}$ generates the
event description $\mathcal{Y}$ conditioned on prompt $\mathcal{P}$. The
kinematic module runs continuously, while the VLM is invoked only for localized
event proposals.

\subsection{Kinematics-Guided Bayesian Inference}
\label{sec:kinematic_bayes}

VIBES tracks each vehicle and decomposes its image-plane motion in a local Frenet frame. For each vehicle, its earlier observations and current neighboring traffic are used to estimate a context-dependent normal-motion distribution, while the recent segment is reserved for anomaly testing. Anomaly-specific Bayesian surprise scores are then computed to generate kinematic triggers when the corresponding probability boundaries are violated.

\subsubsection{Vehicle Trajectory Tracking}
\label{sec:trajectory_tracking}

VIBES combines Slicing Aided Hyper Inference (SAHI)~\cite{akyon2022slicing} with a lightweight multi-object tracker to obtain trajectories for both near- and far-field vehicles. For vehicle $i$ at its $r$-th valid observation, the ground-contact point is denoted by $\mathbf{p}_i^{t_r}\in\mathbb{R}^{2}$, and its image scale is $a_i^{t_r}=\sqrt{b_{w,i}^{t_r}b_{h,i}^{t_r}}$. For two consecutive valid observations, the mean scale is $\bar{a}_i^{t_r}=(a_i^{t_r}+a_i^{t_{r-1}})/2$, and their interval is $\Delta t_r=t_r-t_{r-1}$. The scale-normalized image-plane velocity is $\mathbf{v}_i^{t_r}=(\mathbf{p}_i^{t_r}-\mathbf{p}_i^{t_{r-1}})/(\Delta t_r\,\bar{a}_i^{t_r})$. Here, $\Delta t_r$ accounts for nonconsecutive observations, while scale normalization reduces the dependence of pixel displacement on target distance.

\subsubsection{Kinematic Decoupling via Frenet Frame Formulation}
\label{sec:frenet}

To represent motion consistently across road directions, VIBES constructs a local Frenet frame at each vehicle location. Let $\mathbf{e}_{\parallel,i}^{t_r}$ and $\mathbf{e}_{\perp,i}^{t_r}$ denote the tangent and orthogonal unit vectors of the local road direction. The longitudinal and lateral velocities are $u_i^{t_r}=(\mathbf{e}_{\parallel,i}^{t_r})^{\top}\mathbf{v}_i^{t_r}$ and $w_i^{t_r}=(\mathbf{e}_{\perp,i}^{t_r})^{\top}\mathbf{v}_i^{t_r}$, respectively. The speed magnitude is $s_i^{t_r}=\sqrt{(u_i^{t_r})^2+(w_i^{t_r})^2}$, yielding the kinematic state $\mathbf{k}_i^{t_r}=[u_i^{t_r},w_i^{t_r},s_i^{t_r}]^{\top}$.

At evaluation time $t$, vehicle $i$ has valid observations $t_1<t_2<\cdots<t_{n_i^t}\le t$. Its state sequence $\mathcal{K}_i^t=\{\mathbf{k}_i^{t_r}\}_{r=1}^{n_i^t}$ is divided into an earlier history $\mathcal{H}_i^t=\{\mathbf{k}_i^{t_r}\}_{r=1}^{n_i^t-L_i^t}$ and a recent test segment $\mathcal{R}_i^t=\{\mathbf{k}_i^{t_r}\}_{r=n_i^t-L_i^t+1}^{n_i^t}$, where $L_i^t<n_i^t$. The recent segment is excluded from normal-motion estimation.

The current neighboring observations are $\mathcal{O}_i^t=\{\mathbf{k}_j^t:j\in\mathcal{N}_i^t\}$, where $\mathcal{N}_i^t$ contains vehicles selected by spatial and directional compatibility. Longitudinal inference uses neighbors with compatible travel directions, while speed-based inference may use moving neighbors from either direction.

Finally, $d\in\mathcal{C}=\{\parallel,\perp,\mathrm{spd}\}$ indexes the scalar kinematic components, with $x_{i,\parallel}^t=u_i^t$, $x_{i,\perp}^t=w_i^t$, and $x_{i,\mathrm{spd}}^t=s_i^t$. Accordingly, $\mathcal{H}_{i,d}^t$, $\mathcal{R}_{i,d}^t$, and $\mathcal{O}_{i,d}^t$ denote the corresponding scalar measurements.

\subsubsection{Online Estimation of the Normal-Motion Distribution}
\label{sec:online_normal}

\paragraph{Bayesian formulation of the normal-motion distribution.}
For each vehicle \(i\), scalar kinematic measurement \(d\), and evaluation time \(t\), VIBES estimates a normal-motion distribution from the earlier ego history
\(\mathcal{H}_{i,d}^{t}\) and the current neighboring traffic
\(\mathcal{O}_{i,d}^{t}\).
This distribution specifies the expected value and usual variation of the selected measurement under the current traffic condition.
A recent measurement or motion pattern that falls in a low-probability region of this distribution provides evidence of an anomaly.

The distribution is parameterized by
\(\boldsymbol{\theta}_{i,d}^{t}
=(\mu_{i,d}^{t},(\sigma_{i,d}^{t})^2)\).
The mean \(\mu_{i,d}^{t}\) is the expected value of the selected component, while
\((\sigma_{i,d}^{t})^2\) describes its variation.
For example, \(d=\parallel\) refers to longitudinal velocity,
\(d=\perp\) to lateral velocity, and
\(d=\mathrm{spd}\) to speed magnitude.
These parameters are unknown and may change with the vehicle and traffic context.
To obtain the probability boundary used for anomaly detection, VIBES maintains a probability density over the unknown parameters and updates it when new traffic observations become available.
For an unknown parameter \(\boldsymbol{\theta}\) and observed data \(\mathcal{D}\), Bayes' theorem is
\begin{equation}
p(\boldsymbol{\theta}\mid\mathcal{D})
=
\frac{
p(\mathcal{D}\mid\boldsymbol{\theta})p(\boldsymbol{\theta})
}{
Z(\mathcal{D})
},
\qquad
Z(\mathcal{D})
=
\int
p(\mathcal{D}\mid\boldsymbol{\eta})
p(\boldsymbol{\eta})
\,d\boldsymbol{\eta}
\label{eq:bayes_general}
\end{equation}
Here,
\(p(\boldsymbol{\theta})\) is the prior parameter density,
\(p(\mathcal{D}\mid\boldsymbol{\theta})\) is the likelihood,
\(Z(\mathcal{D})\) is the marginal density of the observations, and
\(p(\boldsymbol{\theta}\mid\mathcal{D})\) is the posterior density.

VIBES uses the earlier ego observations to define the parameter density available before the current neighboring observations are incorporated, and uses the current neighboring observations to define the likelihood.
Assuming that the two data sources are conditionally independent given
\(\boldsymbol{\theta}_{i,d}^{t}\), the posterior density is
\begin{equation}
p\!\left(
\boldsymbol{\theta}_{i,d}^{t}
\mid
\mathcal{H}_{i,d}^{t},
\mathcal{O}_{i,d}^{t}
\right)
=
\frac{
p\!\left(
\mathcal{O}_{i,d}^{t}
\mid
\boldsymbol{\theta}_{i,d}^{t}
\right)
p\!\left(
\boldsymbol{\theta}_{i,d}^{t}
\mid
\mathcal{H}_{i,d}^{t}
\right)
}{
\displaystyle
\int
p\!\left(
\mathcal{O}_{i,d}^{t}
\mid
\boldsymbol{\eta}
\right)
p\!\left(
\boldsymbol{\eta}
\mid
\mathcal{H}_{i,d}^{t}
\right)
\,d\boldsymbol{\eta}
}.
\label{eq:vibes_bayes}
\end{equation}

Each scalar kinematic measurement is modeled by
\(X_{i,d}^{\tau}\mid\boldsymbol{\theta}_{i,d}^{t}
\sim
\mathcal{N}(\mu_{i,d}^{t},(\sigma_{i,d}^{t})^2)\)
The Gaussian mean represents the expected motion, and its variance represents the variation of the measurement.

\paragraph{Derivation of the mean and variance.}

For one scalar kinematic component, we omit the vehicle and component indices and write
$\mathcal{H}^{t}=\{x_{H,1}^{t},\ldots,x_{H,n_H}^{t}\}$ and
$\mathcal{O}^{t}=\{x_{O,1}^{t},\ldots,x_{O,n_O}^{t}\}$.
The history and neighboring observations are modeled as conditionally independent samples around the same expected normal-motion value $\mu_t$:
$X_{H,r}^{t}=\mu_t+\varepsilon_{H,r}^{t}$ with
$\varepsilon_{H,r}^{t}\overset{\mathrm{i.i.d.}}{\sim}\mathcal{N}(0,\sigma_H^2)$, and
$X_{O,j}^{t}=\mu_t+\varepsilon_{O,j}^{t}$ with
$\varepsilon_{O,j}^{t}\overset{\mathrm{i.i.d.}}{\sim}\mathcal{N}(0,\sigma_O^2)$.
The source-specific variances $\sigma_H^2$ and $\sigma_O^2$ are treated as fixed within each update.

For the history and neighboring observations, define the sample means
$\bar{x}_{H}^{t}=n_H^{-1}\sum_{r=1}^{n_H}x_{H,r}^{t}$ and
$\bar{x}_{O}^{t}=n_O^{-1}\sum_{j=1}^{n_O}x_{O,j}^{t}$.
Their sampling variances are
$P_H^{t}=\sigma_H^2/n_H$ and $P_O^{t}=\sigma_O^2/n_O$.
The vehicle history provides an empirical Gaussian prior for the expected
normal motion $\mu^t$, while the current neighbor mean provides new evidence:
\begin{equation}
\begin{aligned}
p_t^{-}(\mu^t)
&=p(\mu^t\mid\mathcal{H}^{t})
=\mathcal{N}(\mu^t;\bar{x}_{H}^{t},P_H^{t}),\\
p(\bar{x}_{O}^{t}\mid\mu^t)
&=\mathcal{N}(\bar{x}_{O}^{t};\mu^t,P_O^{t}).
\end{aligned}
\label{eq:prior_likelihood}
\end{equation}

Combining the history prior with the neighboring evidence yields the Gaussian
posterior
\begin{equation}
p(\mu^t\mid\mathcal{H}^{t},\mathcal{O}^{t})
=
\mathcal{N}(\mu^t;\widehat{\mu}^{t},P^{t}),
\label{eq:mean_posterior}
\end{equation}
where
$P^{t}=((P_H^{t})^{-1}+(P_O^{t})^{-1})^{-1}$ and
$\widehat{\mu}^{t}
=(\lambda_H^{t}\bar{x}_{H}^{t}
+\lambda_O^{t}\bar{x}_{O}^{t})
/(\lambda_H^{t}+\lambda_O^{t})$,
with precisions
$\lambda_H^{t}=n_H/\sigma_H^2$ and
$\lambda_O^{t}=n_O/\sigma_O^2$.
Thus, observations with lower uncertainty contribute more strongly to the
estimated normal-motion mean.

After updating the mean, VIBES estimates the current motion variation from
the combined history and neighboring observations:
\begin{equation}
(\widehat{\sigma}^{t})^2
=
\frac{
\sum_{r=1}^{n_H}(x_{H,r}^{t}-\widehat{\mu}^{t})^2
+
\sum_{j=1}^{n_O}(x_{O,j}^{t}-\widehat{\mu}^{t})^2
}{
n_H+n_O
}
\label{eq:measurement_variance}
\end{equation}
Here, $P^{t}$ represents uncertainty in the estimated mean, whereas
$(\widehat{\sigma}^{t})^2$ characterizes the observed variation of normal
motion under the current traffic context.

\paragraph{Reference distribution for anomaly detection.}
The posterior in Eq.~\eqref{eq:mean_posterior} describes uncertainty in the
expected normal-motion value. Accounting for this uncertainty, a future
normal-motion observation follows the posterior predictive distribution
$\widetilde{X}^{t}\sim
\mathcal{N}(\widehat{\mu}^{t},(\widehat{\sigma}^{t})^2+P^{t})$.
For anomaly testing, VIBES instead uses the plug-in reference distribution
\begin{equation}
q^{t}(x)
=
\mathcal{N}
\left(
x;\widehat{\mu}^{t},(\widehat{\sigma}^{t})^2
\right),
\label{eq:reference_distribution}
\end{equation}
which fixes the expected value at its posterior estimate and defines the
normal-motion boundary using the estimated motion variation. The subsequent
anomaly statistics are evaluated under $q^{t}(x)$.

This estimation is updated online as new observations arrive. At each new
evaluation time $t'$, observations preceding the recent test segment form the
updated history $\mathcal{H}^{t'}$ and refresh the prior for the expected
normal motion. The current neighboring observations $\mathcal{O}^{t'}$ then
provide new evidence to update the posterior mean
$\widehat{\mu}^{t'}$. The combined observations further update the motion
variation $(\widehat{\sigma}^{t'})^2$, yielding a new reference distribution
$q^{t'}(x)$. Consequently, the normal-motion distribution and its probability
boundary adapt continuously to the latest vehicle history and surrounding
traffic.

\subsubsection{Bayesian Surprise Score Formulation}
\label{sec:surprise}

Different anomaly types correspond to different departures from the
context-dependent normal-motion distribution. Lateral anomalies may occur in
either perpendicular direction, whereas overspeeding, deceleration, and
stopping are defined such that larger positive deviations provide stronger
evidence of the corresponding motion pattern.

For a kinematic component $d$, the $L=L_i^t$ recent observations at
$\tau_1<\cdots<\tau_L\le t$ form
$\mathbf{X}_{i,d}^{t}
=[x_{i,d}^{\tau_1},\ldots,x_{i,d}^{\tau_L}]^{\top}$.
The null hypothesis assumes that the recent sequence remains consistent with
the normal-motion distribution estimated from the earlier vehicle history and
current neighboring traffic:
\begin{equation}
H_{0,i,d}^{t}:
\quad
\mathbf{X}_{i,d}^{t}
\mid
\mathcal{H}_{i,d}^{t},
\mathcal{O}_{i,d}^{t}
\sim
\mathcal{N}
\left(
\widehat{\mu}_{i,d}^{t}\mathbf{1}_{L},
\boldsymbol{\Sigma}_{i,d}^{t}
\right)
\label{eq:null_hypothesis}
\end{equation}
For efficient online inference, the recent measurements are treated as
conditionally independent, giving
$\boldsymbol{\Sigma}_{i,d}^{t}
=(\widehat{\sigma}_{i,d}^{t})^2\mathbf{I}_{L}$.

Each anomaly-specific statistic is indexed by
$m\in\{1,\ldots,K\}$, where $d(m)$ denotes the kinematic component associated
with statistic $m$. To express different motion patterns in a unified form,
the recent sequence is centered by its estimated normal-motion mean:
\begin{equation}
\mathbf{R}_{i,d}^{t}
=
\mathbf{X}_{i,d}^{t}
-
\widehat{\mu}_{i,d}^{t}\mathbf{1}_{L},
\qquad
Y_{i,m}^{t}
=
(\mathbf{a}_{i,m}^{t})^{\top}
\mathbf{R}_{i,d(m)}^{t}
\label{eq:projected_statistic}
\end{equation}
The coefficient vector $\mathbf{a}_{i,m}^{t}$ specifies the temporal pattern
captured by each statistic. Equal weights characterize an average motion
deviation, elapsed-time weights capture accumulated lateral deviation, and
contrasting weights over earlier and later observations capture changes in
longitudinal motion.

Under the null hypothesis in Eq.~\eqref{eq:null_hypothesis}, the projected
statistic follows
\begin{equation}
Y_{i,m}^{t}
\mid
H_{0,i,d(m)}^{t}
\sim
\mathcal{N}(0,V_{i,m}^{t}),
\qquad
V_{i,m}^{t}
=
(\mathbf{a}_{i,m}^{t})^{\top}
\boldsymbol{\Sigma}_{i,d(m)}^{t}
\mathbf{a}_{i,m}^{t}
\label{eq:projected_distribution}
\end{equation}
For the observed value $y_{i,m}^{t}$, we define the standardized deviation as
$Z_{i,m}^{t}=y_{i,m}^{t}/\sqrt{V_{i,m}^{t}}$. Under the normal-motion
hypothesis, $Z_{i,m}^{t}$ follows a standard normal distribution.

Let $\alpha\in(0,1)$ denote the per-statistic rejection probability.
The corresponding standard-normal critical values are
$q_{\alpha}^{\mathrm{two}}=\Phi^{-1}(1-\alpha/2)$ for two-sided deviations and
$q_{\alpha}^{\mathrm{one}}=\Phi^{-1}(1-\alpha)$ for one-sided deviations,
where $\Phi$ is the standard-normal cumulative distribution function.
We define the {boundary-normalized Bayesian surprise score} of statistic
$m$ as
\begin{equation}
B_{i,m}^{t}
=
\begin{cases}
|Z_{i,m}^{t}|/q_{\alpha}^{\mathrm{two}},
& \text{either direction},\\
Z_{i,m}^{t}/q_{\alpha}^{\mathrm{one}},
& \text{positive direction},\\
-Z_{i,m}^{t}/q_{\alpha}^{\mathrm{one}},
& \text{negative direction}
\end{cases}
\label{eq:boundary_ratio}
\end{equation}
The score measures the observed kinematic deviation relative to the probability
boundary induced by the current Bayesian normal-motion model. Specifically,
$B_{i,m}^{t}<1$ indicates that the motion remains within the boundary,
$B_{i,m}^{t}=1$ reaches the boundary, and $B_{i,m}^{t}>1$ indicates a boundary
violation. Thus, the score provides a common scale for comparing different
motion statistics whose normal ranges vary with the current traffic context.

\subsubsection{Context-Aware Anomaly Trigger Generation}
\label{sec:trigger}

Each motion statistic produces a boundary-normalized Bayesian surprise score
$B_{i,m}^{t}$. VIBES aggregates these scores at the vehicle level by taking
the largest boundary violation:
\begin{equation}
\begin{aligned}
S_i^t
&=
\max_{1\le m\le K} B_{i,m}^{t},
\qquad
A_i^t
=
\mathbb{I}[S_i^t>1],
\\[-1mm]
A_i^t=1
&\Longrightarrow
t_a=t,
\qquad
m_i^{\star,t}
\in
\arg\max_{1\le m\le K}B_{i,m}^{t}
\end{aligned}
\label{eq:trigger}
\end{equation}
Here, $S_i^t$ is the vehicle-level boundary-violation score, $A_i^t$ indicates
whether an anomaly trigger is generated, and $m_i^{\star,t}$ identifies the
motion pattern with the largest violation. Because each
$B_{i,m}^{t}$ is normalized by its corresponding probability boundary, the
common threshold $S_i^t>1$ indicates that at least one kinematic pattern
deviates beyond its expected normal range.

VIBES considers five representative motion patterns:
\begin{equation}
\begin{aligned}
Y_{i,\mathrm{lat\text{-}v}}^{t}
&=
\bar{w}_i^t-\widehat{\mu}_{i,\perp}^{t},
&
Y_{i,\mathrm{over}}^{t}
&=
\bar{u}_i^t-\widehat{\mu}_{i,\parallel}^{t},
\\
Y_{i,\mathrm{decel}}^{t}
&=
\bar{u}_{i,\mathrm{old}}^t-\bar{u}_{i,\mathrm{new}}^t,
&
Y_{i,\mathrm{stop}}^{t}
&=
\widehat{\mu}_{i,\mathrm{spd}}^{t}-\bar{s}_i^t,
\\
Y_{i,\mathrm{lat\text{-}d}}^{t}
&=
\sum_{r=1}^{L}
\left(
w_i^{\tau_r}-\widehat{\mu}_{i,\perp}^{t}
\right)
\Delta\tau_r
\end{aligned}
\label{eq:kinematic_statistics}
\end{equation}
Here, $\bar{w}_i^t$, $\bar{u}_i^t$, and $\bar{s}_i^t$ are averages over the
recent test segment, while $\bar{u}_{i,\mathrm{old}}^t$ and
$\bar{u}_{i,\mathrm{new}}^t$ denote the longitudinal averages over its earlier
and later parts. The term $\Delta\tau_r$ provides the elapsed-time weight for
the $r$-th recent observation. The lateral-velocity and accumulated lateral
deviation statistics are evaluated in both directions. Overspeeding,
deceleration, and stopping are formulated so that larger positive values
correspond to stronger deviations from their respective normal-motion patterns.

Because $\widehat{\mu}_{i,d}^{t}$ and
$\widehat{\sigma}_{i,d}^{t}$ are updated from the latest vehicle history and
neighboring traffic, the resulting variance $V_{i,m}^{t}$, probability
boundaries, Bayesian surprise scores, and anomaly triggers adapt to the local
traffic condition at each evaluation time.

\subsection{Focused Vision-Language Reasoning}
\label{sec:focused_vlm}

\subsubsection{Spatiotemporal Localization}
\label{sec:localization}

\paragraph{Event proposal formation.}
A frame-level trigger is represented by
$\mathbf{a}_i^t=(t,i,m_i^{\star,t},S_i^t,\mathbf{b}_i^t)$, where $t$ is the
trigger time, $i$ is the vehicle identifier, $m_i^{\star,t}$ denotes the
motion statistic with the largest boundary violation in
Eq.~\eqref{eq:trigger}, $S_i^t=\max_m B_{i,m}^t$ is the corresponding
vehicle-level score, and $\mathbf{b}_i^t$ is the vehicle bounding box.

Triggers that likely correspond to the same traffic event are grouped into
an {event proposal}. Triggers from the same track are grouped when they
occur within a short temporal interval. Triggers with different identifiers
are also merged when they occur at nearby times and occupy the same local
image region, with spatial distance normalized by vehicle scale. This
region-based grouping reduces event fragmentation caused by track-ID changes. Let $\mathcal{A}_e$ denote the trigger set of proposal $e$. Its anchor is the
trigger with the largest vehicle-level score,
$\mathbf{a}_e^{\mathrm{pk}}
=\mathbf{a}_{i_e^{\mathrm{pk}}}^{t_e^{\mathrm{pk}}}
\in\arg\max_{\mathbf{a}_i^t\in\mathcal{A}_e}S_i^t$,
and its associated motion statistic is
$m_e=m_{i_e^{\mathrm{pk}}}^{\star,t_e^{\mathrm{pk}}}$.
The anchor time $t_e^{\mathrm{pk}}$ and vehicle $i_e^{\mathrm{pk}}$ initialize
temporal and spatial localization, while $m_e$ determines the temporal
context required for the event.

\paragraph{Temporal localization.}
Around the anchor, VIBES constructs an event window
$\mathcal{W}_e=[t_e^{\mathrm{pk}}-\tau_e^{-},
t_e^{\mathrm{pk}}+\tau_e^{+}]$ and selects keyframes
$\mathcal{Q}_e=\{\tau_{e,1}<\cdots<\tau_{e,L_e}\}\subseteq\mathcal{W}_e$.
The pre- and post-anchor durations $\tau_e^{-}$ and $\tau_e^{+}$ depend on
the dominant motion pattern, its anomaly score, and the temporal extent of
the proposal. Stopping and deceleration retain more pre-event context to
capture the preceding motion change, whereas strong lateral or longitudinal
deviations may retain additional post-event frames to capture their
subsequent evolution. The selected keyframes summarize the main stages of
the event, including its context, onset, state change or interaction, and
outcome.

\paragraph{Spatial localization and visual input construction.}
For each $\tau\in\mathcal{Q}_e$, VIBES retains the event vehicle and relevant
interacting vehicles. Their bounding boxes define a local region
$\mathbf{B}_e^\tau$, which is enlarged by a margin proportional to the
median vehicle scale $a_k^\tau=\sqrt{b_{w,k}^\tau b_{h,k}^\tau}$.
The resulting focused crop is
$\mathbf{C}_e^\tau=\operatorname{Crop}(\mathbf{I}_\tau,\mathbf{B}_e^\tau)$.

For moving events, the crop is recentered at each selected frame to follow
the event vehicle. For abnormal stopping, a fixed crop centered on the
stopping location is retained across frames, making the motion before and
after stopping directly comparable. The focused crops are arranged in
temporal order to form a storyboard $\mathbf{G}_e$, which preserves the
event progression. A compact comparison image $\mathbf{D}_e$ is constructed
from keyframes around the most evident visual or kinematic transition,
highlighting the decisive local change for VLM inspection.

\subsubsection{Semantic Explanation Generation}
\label{sec:semantic_generation}

For event proposal $e$, semantic reasoning is formulated as
$\mathcal{Y}_e=
\mathcal{F}_{\mathrm{vlm}}(\mathbf{D}_e,\mathbf{G}_e,\mathcal{P})$.
Here, $\mathbf{D}_e$ highlights the key transition, $\mathbf{G}_e$ provides
the temporal event context, and $\mathcal{P}$ instructs the VLM to identify
the event vehicle, determine whether the visual evidence supports an anomaly,
and describe its type or visible consequence.

The VLM is invoked only after kinematic triggering and spatiotemporal
localization. It therefore reasons over compact, event-centered visual
evidence rather than the continuous full-frame stream, reducing irrelevant
visual content and unnecessary computation. If the selected evidence does
not support a clear anomaly, the VLM reports that no definite anomaly is
visible.

\section{Experiments}
\label{sec:experiments}

We conduct experiments on real-world traffic surveillance videos to answer four research questions. \textbf{RQ1} evaluates whether VIBES can detect far-field traffic anomalies and report them within the correct temporal window. \textbf{RQ2} examines whether VIBES can identify the event category and generate accurate fine-grained descriptions. \textbf{RQ3} assesses the computational efficiency of the framework for streaming video processing. \textbf{RQ4} studies the contribution of each component to the overall performance.

\subsection{Datasets}

We evaluate VIBES on two public datasets, TUMTraf~\cite{zhou2025tumtraffic} and Accident~\cite{picek2026accident}, and one independently collected dataset, CPED (Chinese Province Event Dataset). For the public datasets, we retain the videos containing far-field traffic anomalies and manually verify their event windows, categories, and key details. CPED consists of 36 expressway surveillance videos retrieved retrospectively according to manually filed incident reports. For each recording, annotators inspect the complete sequence and mark the target event window, anomaly category, involved vehicles, and key visual and motion characteristics. Compared with the public subsets, CPED contains longer video sequences and is used to evaluate continuous anomaly detection over extended surveillance footage. Table~\ref{tab:datasets} summarizes the dataset statistics.

\begin{table}[t]
    \centering
    \caption{Dataset statistics.}
    \label{tab:datasets}
    \footnotesize
    \setlength{\tabcolsep}{3.6pt}
    \renewcommand{\arraystretch}{0.92}
    \begin{tabular*}{\columnwidth}{@{\extracolsep{\fill}}lccc@{}}
        \toprule
        Feature & TUMTraf & Accident & CPED \\
        \midrule
        Videos & 5 & 16 & 36 \\
        Anomaly Types & 5 & 5 & 4 \\
        Duration (s) & 12--30 & 12--30 & 126--386 \\
        Resolution
        & $960\times600$
        & $960\times600$
        & $1920\times1080$ \\
        FPS & 30 & 30 & 10 \\
        \bottomrule
    \end{tabular*}
\end{table}

\subsection{Experimental Settings}

All experiments are conducted on an Ubuntu 23.04 server equipped with the NVIDIA A40 GPU. Each video is processed as a stream using successive 8-s windows. VIBES performs online Bayesian inference on each frame. When a window produces a high-surprise response, the selected frames and corresponding cropped regions are provided to Qwen3-VL-8B-Instruct~\cite{Qwen3-VL}, which outputs the anomaly status, event category, and event details. All comparison methods use the same temporal windows, prompt, and output format.
\noindent\textbf{Baselines.}
We compare VIBES with eight methods:
1) \textbf{GLM-4.6V}, a general-purpose vision-language model~\cite{glm2025glm46v};
2) \textbf{GLM-5V-Turbo} (\textbf{GLM-5V}), a multimodal model for visual reasoning~\cite{glm2026glm5vturbo};
3) \textbf{Kimi K2.6} (\textbf{KIMI}), a large multimodal agentic model~\cite{kimi2026k26};
4) \textbf{LLaVA-OneVision-2-8B} (\textbf{LLaVA-OV}), a video-capable vision-language model~\cite{an2026llavaonevision2};
5) \textbf{Qwen3-VL-8B-Instruct} (\textbf{Q3-VL-I}), an instruction-tuned multimodal model~\cite{Qwen3-VL};
6) \textbf{Qwen3-VL-8B-Thinking} (\textbf{Q3-VL-T}), a reasoning variant of Qwen3-VL~\cite{Qwen3-VL};
7) \textbf{Qwen3.6-27B} (\textbf{Q3.6}), a larger-capacity Qwen baseline~\cite{qwen2026qwen36}; and
8) \textbf{DeepSCAN}, a training-free coarse-to-fine visual scanning framework~\cite{li2026deepscan}.
The abbreviated names are used in Table~\ref{tab:main_results}.

\noindent\textbf{Evaluation Metrics.}
We use three window-level metrics. The streaming video is divided into successive 8-s windows, and a prediction is counted only when it is produced within the annotated event window. 
1) \textbf{Recall} measures whether the method reports an anomaly in the target window. 
2) \textbf{Event Accuracy} (\textbf{Event Acc.}) requires both a correct anomaly report and the correct event category. 
3) \textbf{Detail Accuracy} (\textbf{Detail Acc.}) further requires a correct description of the key event details, such as the involved vehicles and the distinguishing characteristics of the event. 
All three metrics are binary at the event-window level and are reported as average hit rates over the annotated event windows.

\subsection{Performance Comparison (\textbf{RQ1, RQ2})}

\begin{table*}[t]
    \centering
    \caption{Performance comparison for anomaly detection, event classification, and detail description (\%).}
    \label{tab:main_results}
    \small
    \setlength{\tabcolsep}{3.0pt}
    \renewcommand{\arraystretch}{1.04}
    \begin{tabular*}{\textwidth}{@{\extracolsep{\fill}}llccccccccc@{}}
        \toprule
        Dataset & Metric
        & GLM-4.6V
        & GLM-5V
        & KIMI
        & LLaVA-OV
        & Q3-VL-I
        & Q3-VL-T
        & Q3.6
        & DeepSCAN
        & VIBES \\
        \midrule
        \multirow{3}{*}{TUMTraf}
        & Recall      & 20.00 & 40.00 & 60.00 & 20.00 & 20.00 & 20.00 & 20.00 & 40.00 & \textbf{100.00} \\
        & Event Acc.  & 20.00 & 40.00 & 60.00 & 20.00 & 20.00 & 20.00 & 20.00 & 40.00 & \textbf{80.00} \\
        & Detail Acc. & 20.00 & 40.00 & 60.00 & 0.00  & 20.00 & 20.00 & 20.00 & 40.00 & \textbf{80.00} \\
        \midrule
        \multirow{3}{*}{Accident}
        & Recall      & 25.00 & 37.50 & 87.50 & 12.50 & 18.75 & 18.75 & 18.75 & 25.00 & \textbf{100.00} \\
        & Event Acc.  & 18.75 & 31.25 & 62.50 & 6.25  & 6.25  & 6.25  & 12.50 & 12.50 & \textbf{81.25} \\
        & Detail Acc. & 6.25  & 18.75 & 56.25 & 6.25  & 6.25  & 6.25  & 12.50 & 12.50 & \textbf{75.00} \\
        \midrule
        \multirow{3}{*}{CPED}
        & Recall      & 19.44 & 36.11 & 83.33 & 8.33  & 11.11 & 11.11 & 13.89 & 16.67 & \textbf{94.44} \\
        & Event Acc.  & 13.89 & 27.78 & 77.78 & 2.78  & 5.56  & 5.56  & 8.33  & 11.11 & \textbf{88.89} \\
        & Detail Acc. & 8.33  & 19.44 & 63.89 & 2.78  & 5.56  & 5.56  & 5.56  & 5.56  & \textbf{80.56} \\
        \bottomrule
    \end{tabular*}
\end{table*}

Table~\ref{tab:main_results} reports the average hit rate for each metric over the annotated event windows. VIBES obtains the highest score in every row. Across all evaluated event windows, it achieves aggregate hit rates of 96.49\% for Recall, 85.96\% for Event Acc., and 78.95\% for Detail Acc. These results indicate that the framework is effective not only at triggering an alert within the required window, but also at identifying the event category and describing the evidence needed to interpret the event.

For anomaly detection, VIBES reaches 100.00\% Recall on both TUMTraf and Accident and 94.44\% on CPED. The small decrease on CPED occurs under substantially longer video durations, while the method still detects 34 of the 36 target event windows. KIMI is the strongest baseline, with Recall scores of 60.00\%, 87.50\%, and 83.33\% on the three datasets. VIBES improves these results by 40.00, 12.50, and 11.11 percentage points, respectively. Most direct multimodal models detect fewer than half of the event windows, suggesting that processing the original window alone is often insufficient when the anomalous vehicle is small and the relevant evidence is primarily motion-related.

The semantic results follow the same trend. VIBES attains Event Acc. scores of 80.00\%, 81.25\%, and 88.89\%, together with Detail Acc. scores of 80.00\%, 75.00\%, and 80.56\%. Aggregated over all event windows, it exceeds KIMI by 14.04 percentage points in Event Acc. and 17.54 points in Detail Acc. Detail Acc. is generally lower than Recall because it imposes a stricter requirement: the method should first detect the anomaly and then correctly describe the involved vehicles and the defining event characteristics. The comparatively small gap between the event and detail scores of VIBES shows that the selected evidence usually contains sufficient information for fine-grained interpretation.

VIBES first applies kinematics-guided Bayesian inference to identify unusual motion and then performs focused vision-language analysis on the selected frames and regions. This design narrows the temporal and spatial search range before semantic reasoning. DeepSCAN also improves local visual inspection through coarse-to-fine scanning, but it does not explicitly model temporal motion deviations. The results support the use of motion-based triggering together with focused semantic analysis for far-field traffic events.

\begin{figure}[h]
  \centering
\includegraphics[width=\linewidth]{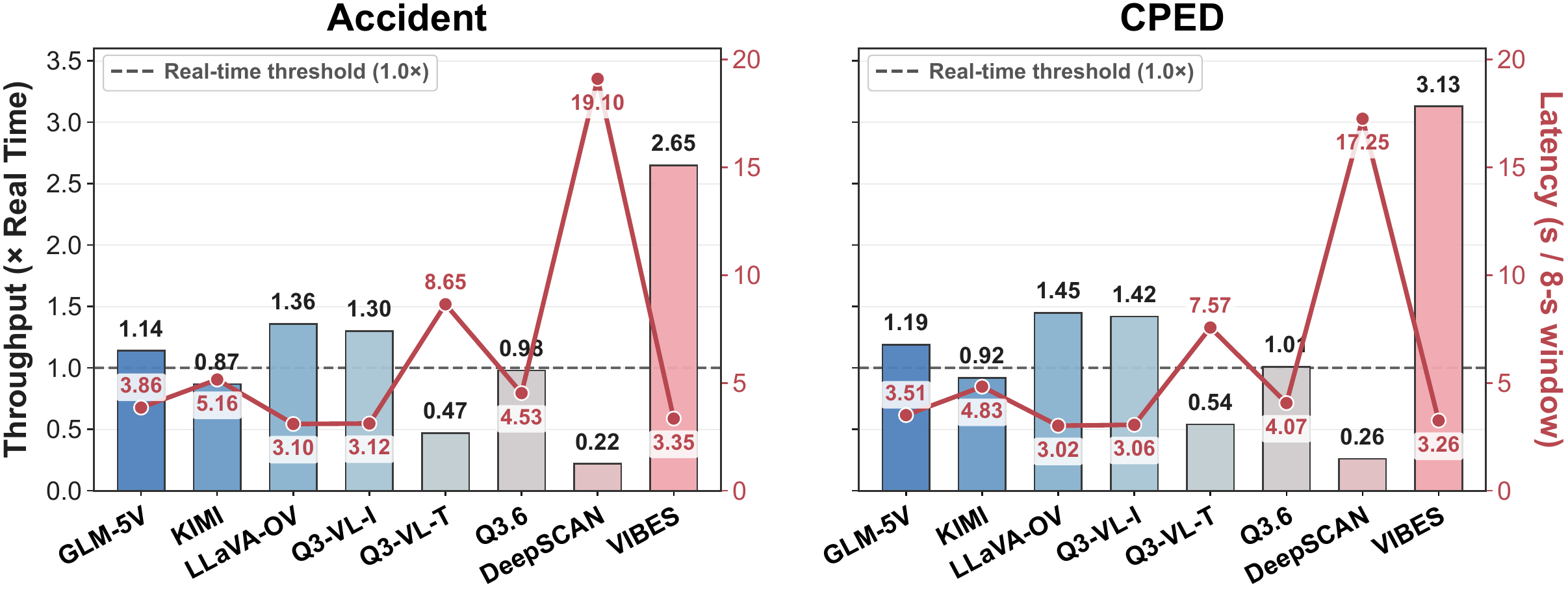}
  \caption{Comparison of computational efficiency.}
  \label{fig:efficiency}
\end{figure}

\subsection{Computational Efficiency Analysis (\textbf{RQ3})}

We evaluate computational efficiency using two complementary metrics, as shown in Figure~\ref{fig:efficiency}. {Throughput} measures stream-level processing speed over the complete video, computed as the ratio of video duration to total processing time; 1.0$\times$ indicates real-time processing. {Latency} measures event-level response time, defined as the elapsed time from feeding an 8-s window containing a target event to obtaining the final textual event explanation.

VIBES achieves the highest throughput on both datasets, reaching 2.65$\times$ on Accident and 3.13$\times$ on CPED, with event-window latencies of 3.35~s and 3.26~s, respectively. LLaVA-OV and Q3-VL-I have slightly lower latency, but their overall throughput and anomaly detection performance are substantially lower, as shown in Table~\ref{tab:main_results}. KIMI and Q3-VL-T remain below the real-time throughput threshold, while DeepSCAN incurs the highest latency and lowest throughput due to repeated visual processing.

The efficiency of VIBES stems from its asynchronous design. The lightweight kinematic module continuously evaluates vehicle motion, while the VLM is invoked only for event proposals triggered by deviations from the dynamically estimated normal-motion distribution. Semantic reasoning is restricted to selected temporal segments and localized visual regions, reducing redundant VLM computation while retaining the evidence required for anomaly detection and explanation.

\label{ablation study}

\begin{figure}[t]
    \centering
    \includegraphics[width=\columnwidth]{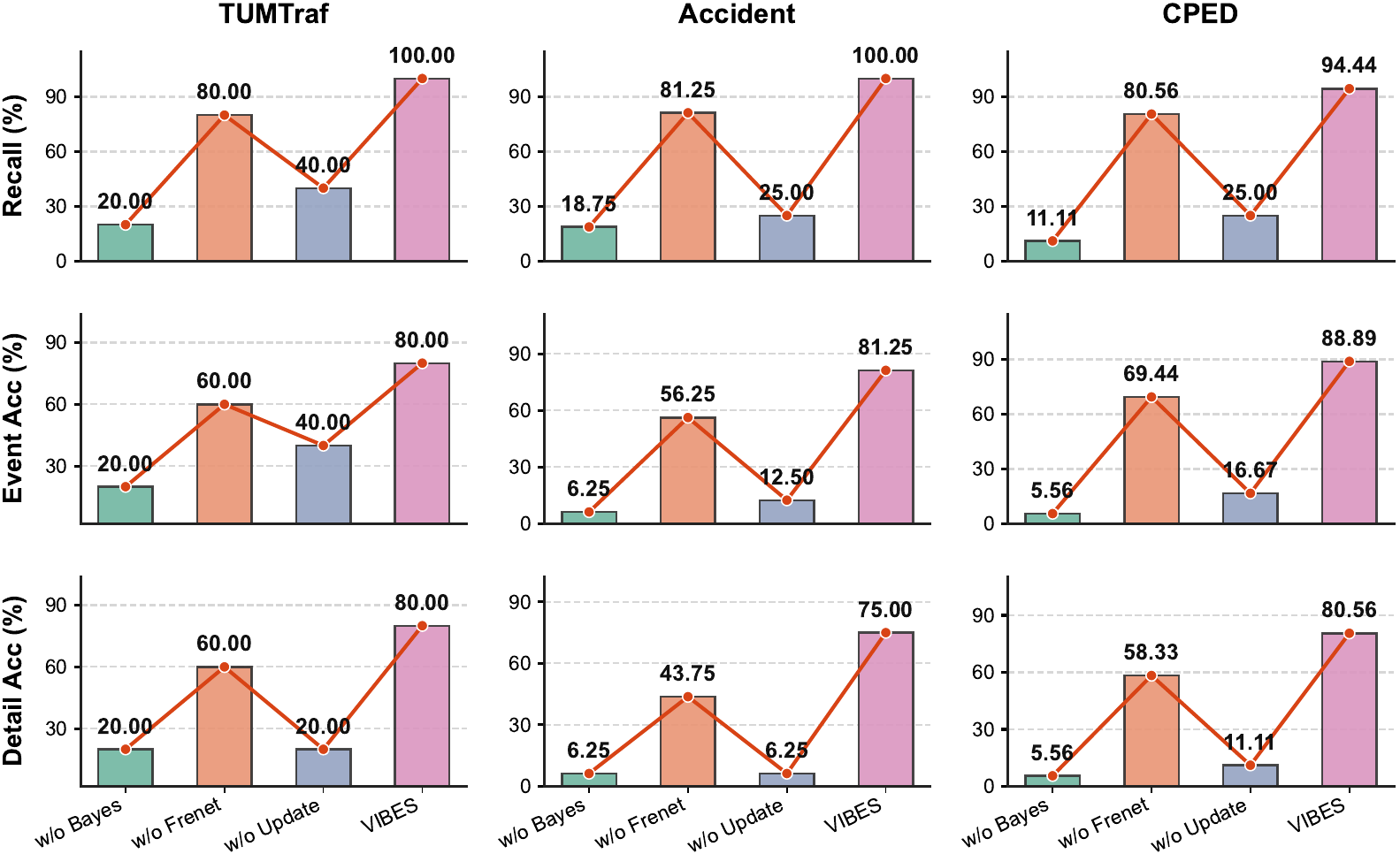}
    \caption{Ablation study.}
    \label{fig:ablation}
\end{figure}

\begin{figure*}
    \centering
    \includegraphics[width=\textwidth]{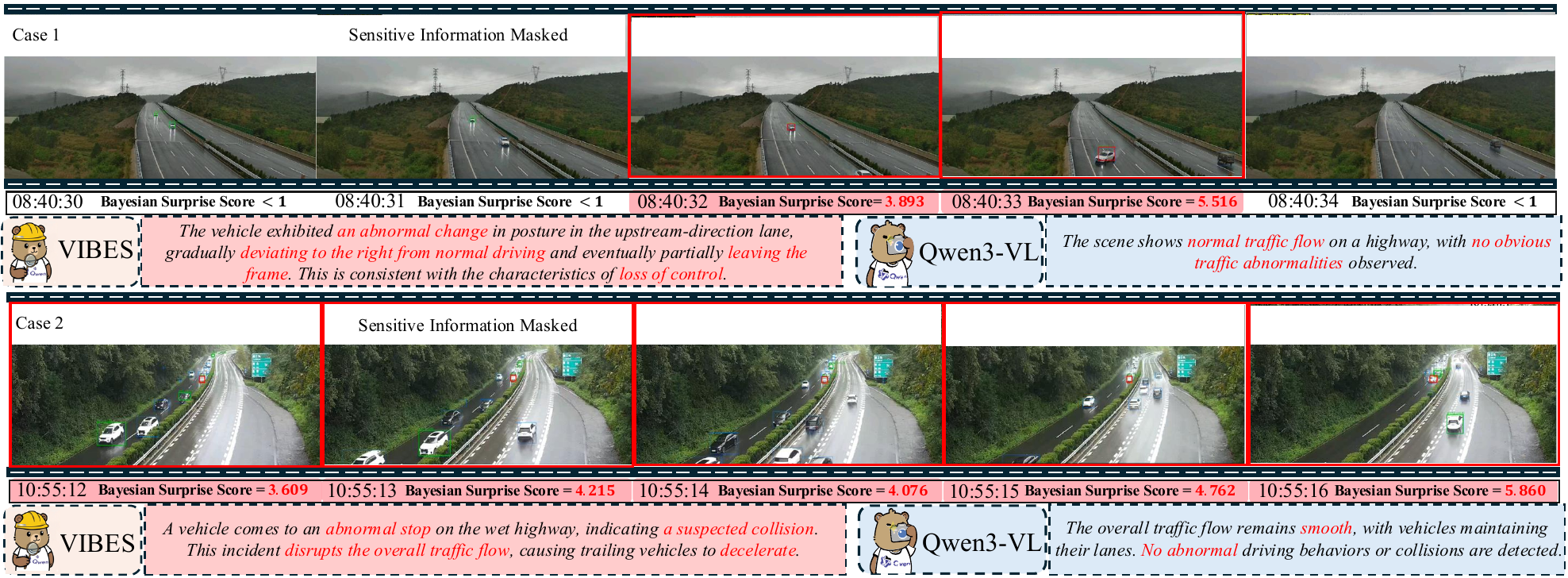}
    \caption{Case study on real-world expressway surveillance videos from CPED. }
    
    \label{fig:case_study}
\end{figure*}


\subsection{Ablation Study (RQ4)}

To evaluate the contribution of each component, we compare VIBES with three variants on the TUMTraf, Accident, and CPED datasets (Fig.~\ref{fig:ablation}). \textbf{1) w/o Bayes} removes the kinematics-guided Bayesian inference and directly applies the VLM to global frames without spatiotemporal localization. The clear performance drop demonstrates the importance of motion-based localization for far-field anomaly detection. \textbf{2) w/o Frenet} replaces the local Frenet representation with absolute image-plane motion, leading to consistent degradation and demonstrating the benefit of decoupling longitudinal and lateral motion across different road directions. \textbf{3) w/o Update} uses static probabilistic boundaries instead of online updating. Its reduced performance shows that adapting the normal-motion distribution to current vehicle history and surrounding traffic is important under changing traffic conditions.

\subsection{Case Study (RQ1, RQ2)}

Figure~\ref{fig:case_study} presents two representative far-field anomalies under different traffic conditions. In Case 1, a distant vehicle exhibits a sudden lateral deviation within a short interval and soon moves out of the camera view. Such brief far-field events are easy to miss in full-frame analysis, yet timely detection is important for prompt incident response. VIBES captures the abnormal lateral motion through its kinematic statistics and localizes the corresponding temporal segment and vehicle region, providing focused evidence for subsequent VLM reasoning.

Case 2 shows a denser traffic scenario in which the target vehicle decelerates and eventually enters an abnormal stopping state. By combining the vehicle's preceding motion with current neighboring traffic, the online Bayesian model identifies these deviations through elevated Bayesian surprise scores and triggers the relevant spatiotemporal region. Focused VLM reasoning over the localized evidence further indicates a suspected collision and its effect on surrounding traffic. In contrast, direct VLM reasoning over the full frames fails to identify the distant abnormal event. Together, the two cases show that motion-guided localization helps preserve subtle and transient far-field evidence for semantic interpretation.

\section{Conclusion}

We proposed VIBES, an asynchronous framework for far-field anomaly detection in expressway surveillance videos. VIBES combines lightweight vehicle kinematics with online Bayesian inference to update context-dependent normal-motion boundaries and trigger focused spatiotemporal localization. The localized visual evidence is then processed by the VLM for semantic reasoning, reducing irrelevant content and unnecessary VLM invocations. Experiments show that VIBES improves far-field anomaly detection while maintaining real-time efficiency across diverse expressway conditions.

\begin{acks}
This work was supported by Beijing Natural Science Foundation (No. L252034).
\end{acks}

\bibliographystyle{ACM-Reference-Format}
\balance
\bibliography{ref}

\end{document}